\documentclass[authoryear,3p,12pt]{elsarticle}

\usepackage{graphics}
\usepackage{epsfig}
\usepackage{graphicx}

\usepackage{amssymb}
\usepackage{amsmath}
\usepackage{subfigure}
\usepackage{color}
\usepackage{setspace}
\usepackage{rotating}
\usepackage{url}
\usepackage{gensymb}
\usepackage{floatrow}
\usepackage{multirow}
\usepackage{float}
\usepackage{pdflscape}

\usepackage[normalem]{ulem} 
\usepackage{smartdiagram}

\usepackage{array}
\newcolumntype{S}{>{\centering\arraybackslash}m{0.6cm}}
\newcolumntype{M}{>{\centering\arraybackslash}m{2.4cm}}
\newcolumntype{L}{>{\centering\arraybackslash}m{8.5cm}}

\usepackage{lineno}

\definecolor{darkgreen}{rgb}{0,.75,0}

\newif\ifsqueezing
\newif\ifaggressivesqueezing

\ifsqueezing
\ifaggressivesqueezing
\newcommand{\Caption}[1]{\caption{{\footnotesize #1}}}

\else
\newcommand{\Caption}[1]{\caption{#1}}

\fi
\else

\newcommand{\Caption}[1]{\caption{#1}}

\fi

\usepackage{color}

\DeclareGraphicsExtensions{.png}

\journal{ISPRS Journal of Photogrammetry and Remote Sensing~}

\begin{document}

\begin{frontmatter}



\title{Geocoding of trees from street addresses and street-level images}


\author[label1]{Daniel~Laumer}
\author[label1]{Nico~Lang}
\author[label3]{Natalie~van~Doorn}
\author[label4]{Oisin~Mac~Aodha}
\author[label2]{Pietro~Perona}
\author[label1]{Jan~Dirk~Wegner}

\address[label1]{EcoVision Lab, Photogrammetry and Remote Sensing, ETH Z\"urich, Switzerland}
\address[label3]{Forest Service, US Department of Agriculture, USA}
\address[label4]{University of Edinburgh, UK}
\address[label2]{California Institute of Technology, USA}


\begin{abstract}
We introduce an approach for updating older tree inventories with geographic coordinates using street-level panorama images and a global optimization framework for tree instance matching. Geolocations of trees in inventories until the early 2000s where recorded using street addresses whereas newer inventories use GPS. Our method retrofits older inventories with geographic coordinates to allow connecting them with newer inventories to facilitate long-term studies on tree mortality etc. What makes this problem challenging is the different number of trees per street address, the heterogeneous appearance of different tree instances in the images, ambiguous tree positions if viewed from multiple images and occlusions. To solve this assignment problem, we (i) detect trees in Google street-view panoramas using deep learning, (ii) combine multi-view detections per tree into a single representation, (iii) and match detected trees with given trees per street address with a global optimization approach. Experiments for $>50000$ trees in 5 cities in California, USA, show that we are able to assign geographic coordinates to 38 \% of the street trees, which is a good starting point for long-term studies on the ecosystem services value of street trees at large scale.

\end{abstract}


\end{frontmatter}

\section{Introduction}\label{sec:intro}

An urban forest inventory represents a “snapshot in time” of trees in a given geographic area \citep{roman2013}. Urban tree inventories are gaining in popularity and increasingly entering the digital age, with digital devices replacing index cards, and data posted on websites facilitating public access and management \citep{cumming2008}. Of particular interest to municipalities are street tree inventories since street trees lie within their management scope and are a publicly visible component of the urban forest. Street tree inventories may be used for directing management priorities and assessing pest and pathogen risk \citep{bond2013}, surveying species diversity and size distribution \citep{mcpherson2013}, and estimating ecosystem services \citep{mcpherson2005,mcpherson2016}. When updated over time, inventories can become key components of long-term monitoring data, which are essential for understanding changes in urban forests \citep{roman2013}. Repeated inventories have been used by researchers to explore trends in street tree populations such as changes in species composition \citep{dawson1985}, tree health impacts of storms \citep{hallet2018}, estimates of demographic rates \citep{roman2013,roman2014,widney2016}, as well as determinants of mortality or growth \citep{koeser2013,vandoorn2018}. Analyses of repeated tree measurements may help predicting population stability and identify trends. Baseline estimates of demographic rates such as mortality or replacement rates provide a basis for comparison to future measurements. Having information on how the urban forest is changing allows managers to identify the more vulnerable segments of the tree population and to focus management efforts. For example, areas of higher mortality can imply a need for maintenance, increased replacement planting, or a change in the species palette. Spatially-explicit tree data are important for several reasons \citep{alonzo2016}: ecosystem functions may vary throughout a city due to legacy effects \citep{roman2017}, species may have different threats from pests, diseases, and fire \citep{santamour1990,lacan2008}, and distribution of ecosystem services may be contributing to environmental injustices \citep{landry2009}. Geographic coordinates are key pieces of information for long-term monitoring studies because they allow for the same tree to be tracked and measured through time with some degree of certainty \citep{vandoorn2018b}.


For modern inventory collection, tress are typically mapped with GPS devices and thus come with a fairly accurate geographic coordinate, making repeated measurements much easier. However, before the prevalence and ease of GPS, trees in inventories were typically only referenced by street address. Manually linking such tree inventories to GPS-measured inventories is infeasible at a large scale. As a result, a lot of valuable inventory information remains unexplored because there is no automated way, yet, to link those two types of inventories. 

Our main motivation is to provide a tool for building long-term tree data sets that can be used to assess changes in the urban forest and corresponding changes in ecosystem services. Although street trees make up a small proportion of the urban forest, they provide considerable ecosystem services~\citep{mcpherson2016}, and disservices to the urban landscape ~\citep{escobedo2011,pataki2011}. Benefits include improvement in air quality, a reduction of the heat island effect, increased carbon capture and storage, rising property values, and an improvement in individual and community wellbeing~\citep{nowak2002,mcpherson2016}. According to the most recent estimate~\citep{mcpherson2016} there are 9.1 million trees lining the streets of California with an ecosystem services value of \$1 billion per year or \$111 per tree, i.e. \$29 per inhabitant of California. However,  little information is available on how street tree populations are changing in terms of demographic rates. While static data sets such as canopy cover and plot data can provide a snapshot in time, demographic details such as growth rates and mortality rates, as well as changes in tree health through time require assessment of the same trees more than once. Many municipalities do not have the funding to complete repeat inventories and/or may place little priority on matching individual trees to create a time series. Thus, making accessible the information in legacy inventories is of high interest to both researchers interested in quantifying tree demographic rates and ecosystem services and for urban forest practitioners interested in data-driven management (e.g., knowing where there are hotspots of tree mortality, or high growth). 

The hope is that our approach on retrofitting existing street tree inventories with geographic coordinates will enable large-scale longitudinal studies, where data of the same population of trees can be analysed over decades. Emphasis of our approach is thus on assigning geographic coordinates to a high absolute number of trees across the state of California as opposed to geocoding each individual tree of the given data base correctly. 

In order to geo-code hundreds of thousands of street trees across California with corresponding addresses, we analyse publicly available street-level panoramas across the whole state. For this purpose, we propose to start from a simplified, computationally more efficient version of~\citep{wegner2016,branson2018} for tree detection and geo-localization. We replace the complex and computationally costly conditional random field formulation with a weighted average that condenses multiple detections of the same tree from different views into a single one. Tree instances detected in images are matched to inventory entries that come with street addresses. 
%
%
Preparing and training the system basically consists of manually labeling trees in images (four days), training the tree detector (four hours), and restructuring all inventories into a homogeneous format (one day). Our system comes at virtually no cost if we put aside costs for running computers and downloading Google images. Given appropriate hardware and a fast internet connection for downloading images, the method scales to arbitrarily large data sets.  


\section{Related Work}\label{sec:related}

{\bf Municipial tree inventories} are traditionally collected by urban forestry professionals (either municipal staff or contracted-out tree management company arborists) but more recently, some programs also have significant citizen science components \citep{cozad2005,bloniarz1996,roman2017}. Variables collected in tree inventories vary by program goal, but the common minimum dataset typically includes tree species, condition, and location \citep{ostberg2013,vandoorn2018b}. 
%
The most common method to record tree location is by street address \citep{roman2013}. Street addresses, while a viable alternative when no technology is available \citep{vandoorn2018b}, has known drawbacks \citep{crown2018}. For example, NYC Parks found that the street address method lacked precision, and created confusion at large building sites where multiple trees were linked to one address. In 2005, NYC Parks made an effort to switch from street addresses to geocoded locations, but 23000 locations could not be geocoded because of incorrect location data \citep{crown2018}, suggesting that the link between the old datasets and new ones cannot be made, resulting in a break in continuity of the dataset. 

Cities and communities interested in maintaining their tree inventories are tasked with converting approximate tree locations from older inventories to more accurate locations from modern geocoding methods. However, as methods of data collection are refined, there is a risk that older inventories without accurate location data may be unusable for urban tree monitoring. One approach to translating past municipal inventories includes georeferencing via GPS and on-screen Geographic Information Systems, as exemplified in Ithaca, NY \citep{city2018}. This approach may be feasible for smaller cities but may be too costly and time consuming for larger cities with more trees. 

{\bf Tree detection} in an automated way, especially in forests, has attracted much attention in the recent 30 years (see~\citet{larsen2011} and~\citet{kaartinen2012} for a detailed comparison of methods until 2012). More recent works usually rely on airborne LiDAR~\citep{lahivaara2014,zhang2014,kwak2017,matasci2018}, a combination of LiDAR and aerial imagery~\citep{qin2014,paris2015}, only aerial imagery~\citep{yang2009}, high-resolution satellite imagery~\citep{li2016}, or medium resolution satellite imagery where tree identification and counting is cast as a semantic density estimation problem~\citep{rodriguez2018}. Automated detection of trees in cities has received less attention.~\citet{straub2003} designed a hierarchical workflow by first segmenting aerial images and height models into consistent regions at multiple scales and then refining tree crown shapes with active contours.~\citet{lafarge2012} produced a seminal work in the urban environment, creating 3D city models from dense aerial LiDAR  point clouds. Besides trees, the authors reconstructed buildings and the ground. After an initial semantic segmentation with a breakline-preserving markov random field (MRF), 3D templates consisting of a cylindrical trunk and an ellipsoidal crown were fitted to the data points.~\citet{jaakkola2010} also modeled tree trunks as cylinders with LiDAR point clouds acquired with an unmanned aerial vehicle (UAV) whereas~\citet{monnier2012} applied a similar approach to LiDAR data acquired with a moving vehicle on the ground. For completeness we note that there is a lot more research on vegetation species mapping but we are not dealing with it in this work and thus a full review is beyond the scope of this paper. We kindly refer the reader to~\citep{alonzo2014,wegner2016,liu2017,branson2018,hartling2019,aval2019,sidike2019} for some of the most recent works.

{\bf Urban object detection} from ground-level imagery is a core research topic in computer vision often approached from an autonomous driving perspective with various existing public benchmarks like KITTI~\cite{geiger2013vision}, CityScapes~\cite{cordts2016cityscapes}, or Mapillary~\cite{neuhold2017mapillary}. In these scenarios, dense image sequences are acquired with minor viewpoint changes in driving direction with forward facing cameras enabling object detection and re-identification across consecutive views~\cite{chen2017multi,zhao2018object}. In contrast, we work with existing Google street-view panorama images that come with wide baselines (usually around 50 meters) between image acquisition locations, image stitching artefacts, only coarse, inaccurate pose information, and strong scale change between views. In recent work~\citep{nassar2019jurse}, we propose a siamese CNN approach that combines visual cues and coarse pose estimation of pairs of images to re-identify objects in street-level images. This work was extended \citep{nassar2019iccv} to a full multi-view object detection approach that learns urban object detection end-to-end using visual appearance cues in images and coarse pose evidence.

Other methods geolocalize ground level images using different techniques to match features to corresponding ground-level images database~\citep{Lin_2013_CVPR,muller2018geolocation}. For example, ground-level imagery can be combined with aerial imagery to accomplish a pixel-accurate semantic segmentation of the scene, or road respectively~\citep{zhai2017predicting,mattyus2016hd}.~\citet{lefevre2017toward} detects changes by with siamese CNNs through comparing aerial imagery and street view panoramas warped to aerial image geometry. \citet{cao2018integrating} classify land use categories in urban areas with semantic features from both aerial and ground-level images. \citet{krylov2018automatic} geo-localize traffic lights and telegraph poles in street-level images with monocular depth estimation using CNNs. A Markov Random Field triangulates object positions. The same authors extend their approach by adding LiDAR data for object segmentation, triangulation, and monocular depth estimation for traffic lights~\cite{krylov2018object}. \citet{zhang2018using} propose a CNN-based object detector for poles and apply a line-of-bearing method to estimate the geographic object position. In this paper, we advocate for a simplified version of~\citep{wegner2016,branson2018} for tree detection and geo-localization that is less costly to compute then a full end-to-end approach \citep{nassar2019iccv} at very large scale (i.e., 48 cities in California.


\section{Methods}\label{sec:method}
\begin{figure}[!ht]
	\centering
	\includegraphics[width=1.0\linewidth]{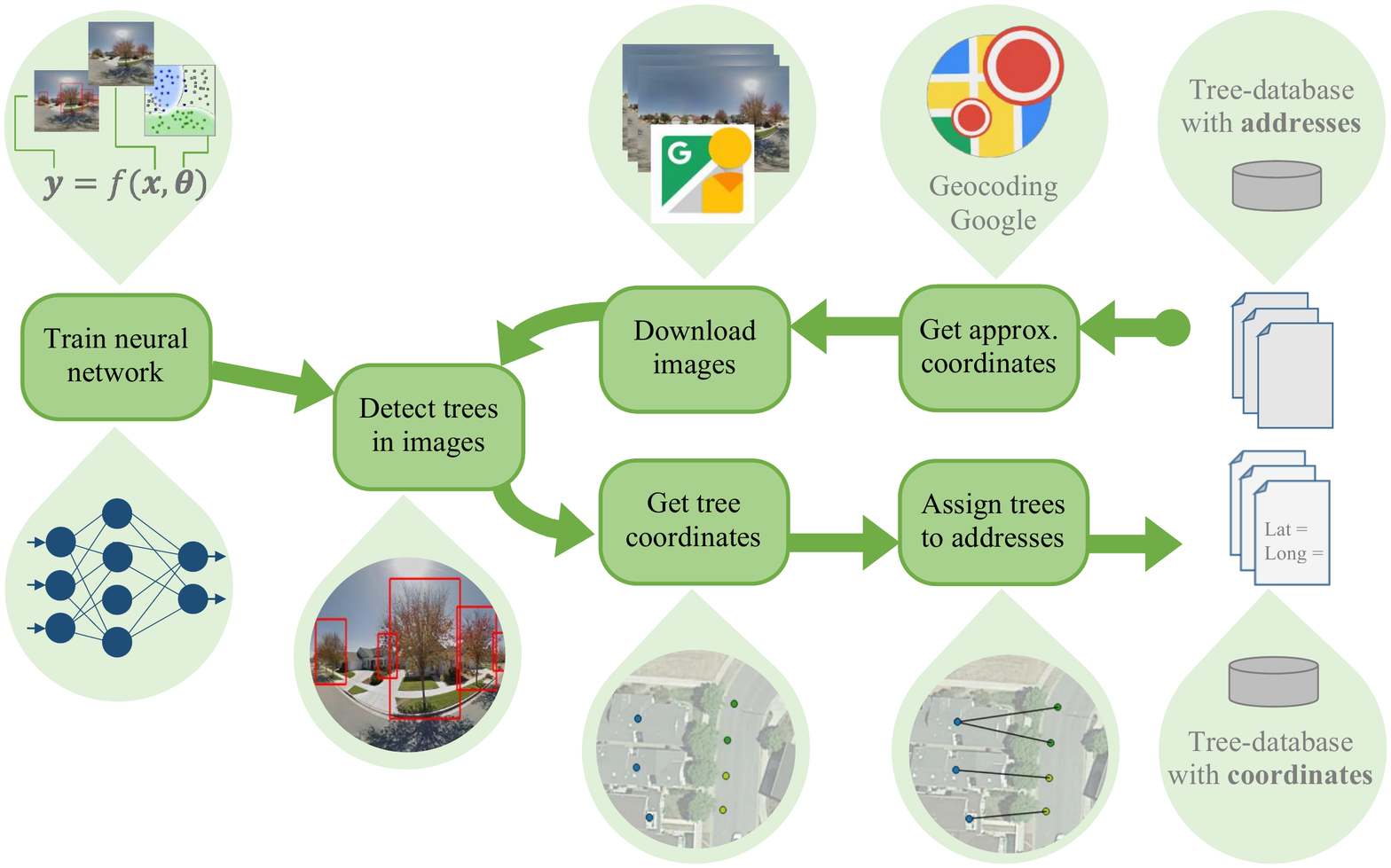} \\
	\Caption{Overview of the workflow for assigning geographic coordinates to trees of an inventory data base given street addresses and street-level panorama images.}
	\label{fig:flowchart}
\end{figure}
In this section, we describe our method to assign geographic coordinates to street trees given street addresses and ground-level panorama images. 
%
%
Given a data base of trees with street addresses, our new system presented here does the following (see flowchart in Fig.~\ref{fig:flowchart}):
\begin{itemize}
	\item It retrieves approximate geographic coordinates for each address,
	\item downloads the available street-view panorama images for each address,
	\item automatically detects all trees per image,
	\item integrates individual tree detections per image into a single geographic coordinate per tree,
	\item and finally matches each detected tree with an entry in the data base. 
\end{itemize}
Because tree assignments to street addresses can be ambiguous due to multiple trees per address as well as trees that can be assigned to more than one address, we solve tree matching via an optimization  framework. Geo-coded trees detected from images are matched to tree entries in the data base with geo-coded street addresses. By globally minimizing the sum over the misalignment between geographic coordinates of detected trees and geographic coordinates of street addresses of trees in the inventory, we seek a globally optimal solution per city. In order to avoid having to match all possible pairs of detected trees and tree entries per inventory, we introduce simplifying assumptions that take into account the locally constrained nature of the matching problem. 




\subsection{Geocoding of street addresses} \label{sec:geocoding}

We reformat all street addresses collected manually in situ into a unified representation that can be processed by the Google geocoding API. 
Approximate coordinates of street addresses are generated using a Google API\footnote{\url{https://developers.google.com/maps/documentation/geocoding/intro}}, which returns geo-coordinates centered on the building of a particular parcel (i.e., portion of land attributed to a certain street address).
In order to detect outliers before further processing (i.e., addresses projected far away from the rest of the coordinates), we rely on the z-score for latitude $z_{i,lat}$ and longitude $z_{i,long}$. The z-score compares the position of each individual latitude $lat_{i}$ and longitude $long_{i}$ to the average values $\mu_{lat}$ and $\mu_{long}$ across all coordinates $i=1,2,3,...,n$ (with $n$ being the total number of street addresses in the whole set) in terms of standard deviations $\sigma_{lat}$ and $\sigma_{long}$ (Eq.~\ref{eq:zscore}). Note that in our case, a set is equal to a tree inventory of an individual municipality and outliers are thus tree positions projected beyond the boundaries of a municipality.  

\begin{equation}\label{eq:zscore}
z_{i,lat} = \frac{lat_i - \mu_{lat}}{\sigma_{lat}}, ~~z_{i,long} = \frac{long_i - \mu_{long}}{\sigma_{long}}
\end{equation}
with
\begin{equation}
\mu_{lat} = \frac{1}{n} \sum_{i = 1}^{n} lat_i, ~~\sigma_{lat} = \sqrt{ \frac{1}{n} \sum_{i = 1}^{n} (lat_i - \mu_{lat})^2}
\end{equation}
\begin{center}
outlier if $z_{i,lat} > 3$ OR $z_{i,long} > 3$\\
\end{center}

All points with a z-score above a certain threshold are discarded. We empirically found that coordinates further away from the set than three times the standard deviation are outliers and thus set the threshold to 3. 

It should be noted that the Google geocoding API uses different ways and accuracy levels to geocode street addresses. In 68\% of our cases, it returns geo-coordinates that are located in the center of the biggest building on the parcel at a particular street address. For all remaining cases where rooftop coordinates are unavailable, less accurate approximations are made. Approximations are usually computed by interpolating between junctions or along a polyline or region. Resulting coordinates are often located on the road in front of the parcel instead of being positioned at the center of the biggest building. It turns out that these cases complicate the matching process, because the geocoded coordinates of street addresses on either road side are located in very close proximity leading to a higher risk of tree and address mismatch. We investigate how this affects our results in the experimental section.

\subsection{Tree detection}

\begin{figure}[!ht]
	\centering
	\includegraphics[width=1.0\linewidth]{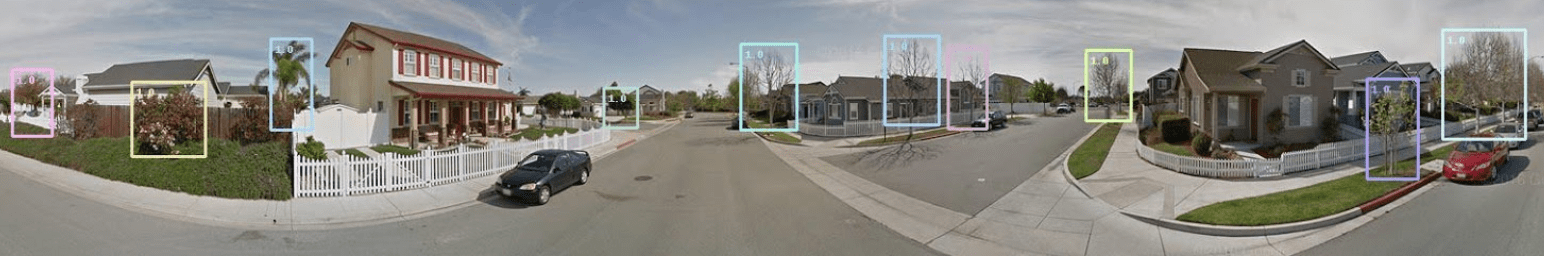} \\
	\Caption{Bounding boxes around automatically detected trees in the center slice of an example street-view panorama. (Imagery \textcopyright~2019 Google)}
	\label{fig:bboxexample}
\end{figure}

For tree detection, we download Google street-view panoramas within a 50 meter radius around each address coordinate, detect all trees in the panoramas, project each individual detection to geo-coordinates and combine all individual, geocoded detections per tree into a single geographic location. Trees in street-view panorama images are detected with the Faster R-CNN method~\citep{renNIPS15fasterrcnn}, an object detector using deep convolutional neural networks which we already applied successfully in our recent works~\citep{wegner2016,branson2018}. The output of the tree detector is one bounding box per detected tree in each street-view panorama (see example in Fig.~\ref{fig:bboxexample}). 
We project all bounding boxes of detected trees per panorama to geographic coordinates assuming locally flat terrain with the equations given in~\citet{wegner2016} and~\citet{branson2018}. More precisely, we assume that the terrain is locally flat and that the camera is levelled against the ground (pitch = 0). This assumption implies that the horizon is always in the horizontal center-line of the panorama. We empirically found that this simplifying assumption holds for most cases except roads along high slopes, where trees on the road side are positioned either above or below the road surface. Another exception is road surfaces with high degrees of tilt. However, we did not face any of these complex scenarios in our data set because it does not include municipalities in steep mountainous areas. 

In order to compute the position of a tree in geo-coordinates from a single ground-level panorama image, we need to estimate its distance from the given camera position and its direction (camera position and heading come with the panorama meta-file). We take the vertical center axis of a detected bounding box as reference for the direction. With an average camera height above ground of 3 meters, we project each individual detection to geo-coordinates with the center point of the bottom line per bounding box. Projection of tree bounding boxes to geo-coordinates is performed for each detected tree per image. With an average spacing of 15 meters between panoramas, each tree is usually detected multiple times. 

We propose a faster to compute, less complex, and easily scalable method of our original tree detection method ~\citep{wegner2016,branson2018}. The conditional random field of~\citep{wegner2016,branson2018} is replaced with a simple weighted average. More precisely, we compute a score per tree detection by inverse distance weighting of the object detection scores. We select the detection with the highest score per neighborhood and sample all tree detections falling within a four meter radius around it. All coordinates of this set of tree detections are combined in a weighted average with the weight being the distance weighted scores. 

\begin{figure}[!ht]
	\centering
	\begin{tabular}{ccc}
		\includegraphics[width=0.29\linewidth]{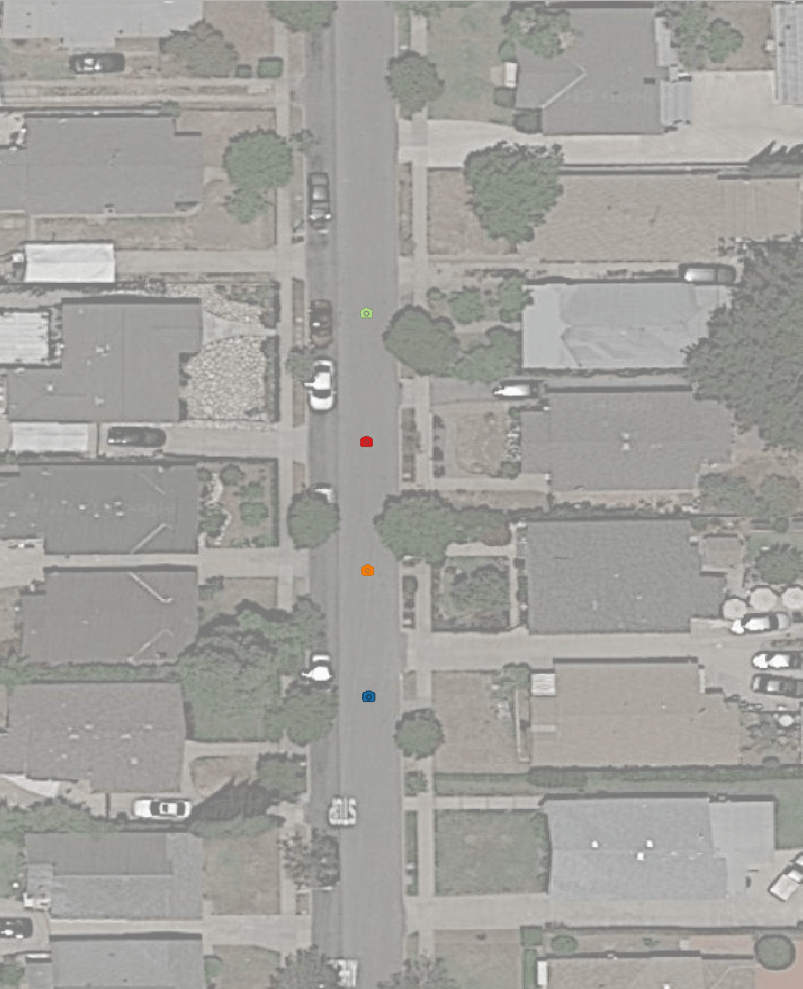}& 
		\includegraphics[width=0.29\linewidth]{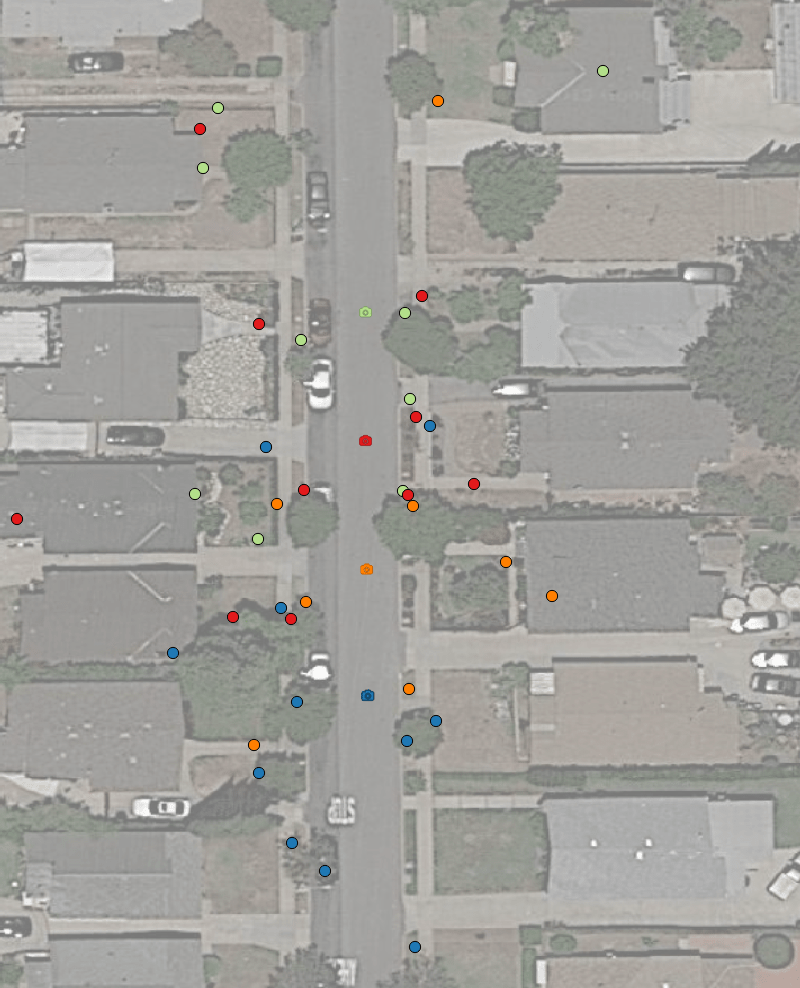}&
		\includegraphics[width=0.29\linewidth]{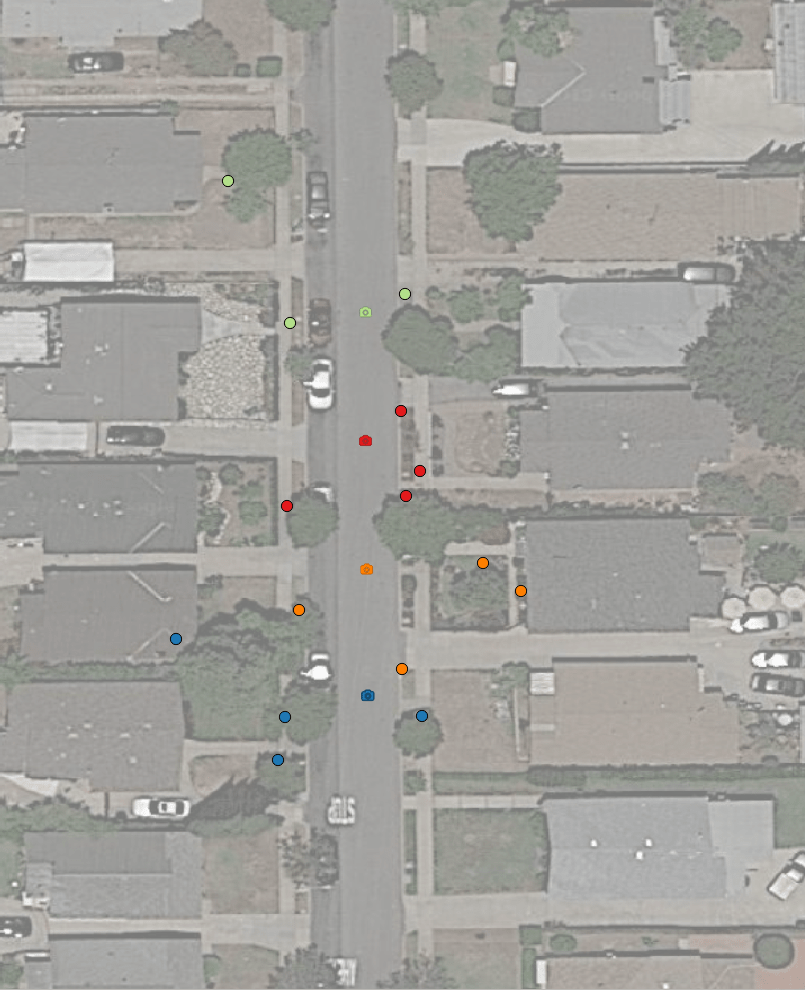}\\
		(a) & (b) & (c)\\
	\end{tabular}
	\Caption{Geocoding of tree detections with inverse distance weighting of detection scores combining multiple detections per individual tree into a single position in geographic coordinates. (a) positions of street-view cameras, (b) all tree detections projected to geographic positions (color encodes camera), (c) final tree positions after non-maximum suppression. (Imagery \textcopyright~2019 Google)
	}
	\label{fig:nonmaxsuppr}
\end{figure}

\begin{figure}[!ht]
	\centering
	\begin{tabular}{cc}
		\includegraphics[width=0.45\linewidth]{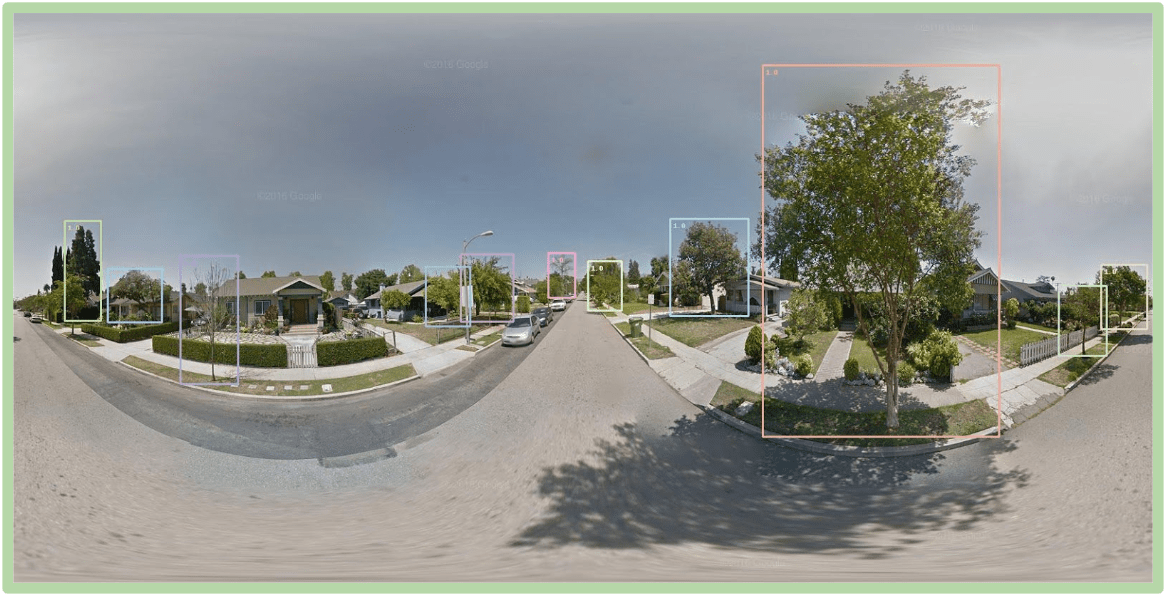}& 
		\includegraphics[width=0.45\linewidth]{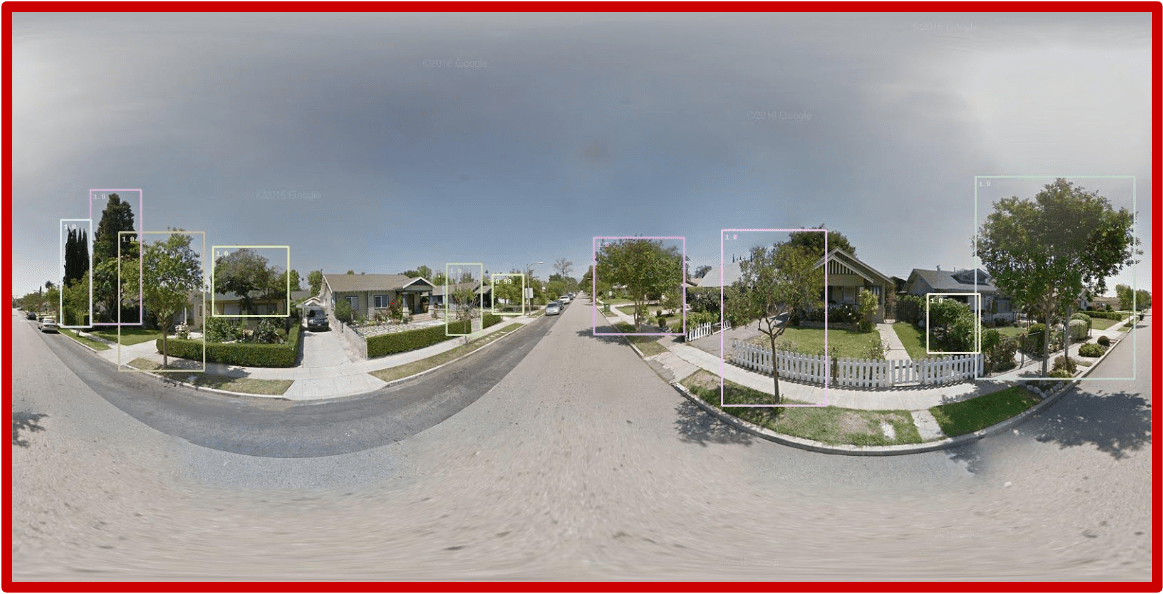}\\
		(a) & (b)\\
		\includegraphics[width=0.45\linewidth]{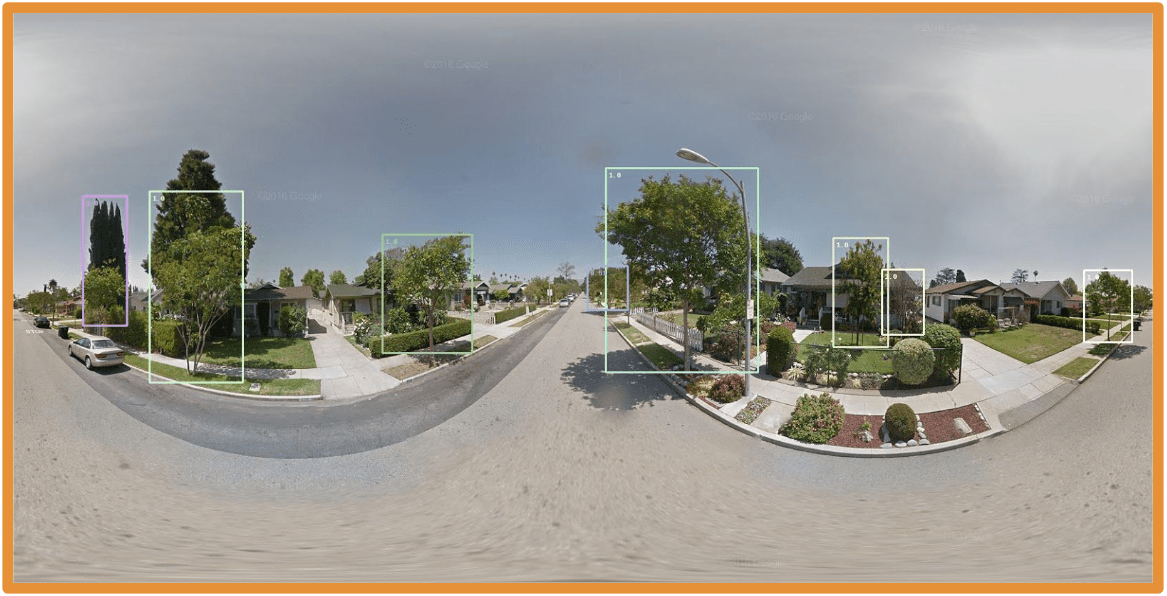}& 
		\includegraphics[width=0.45\linewidth]{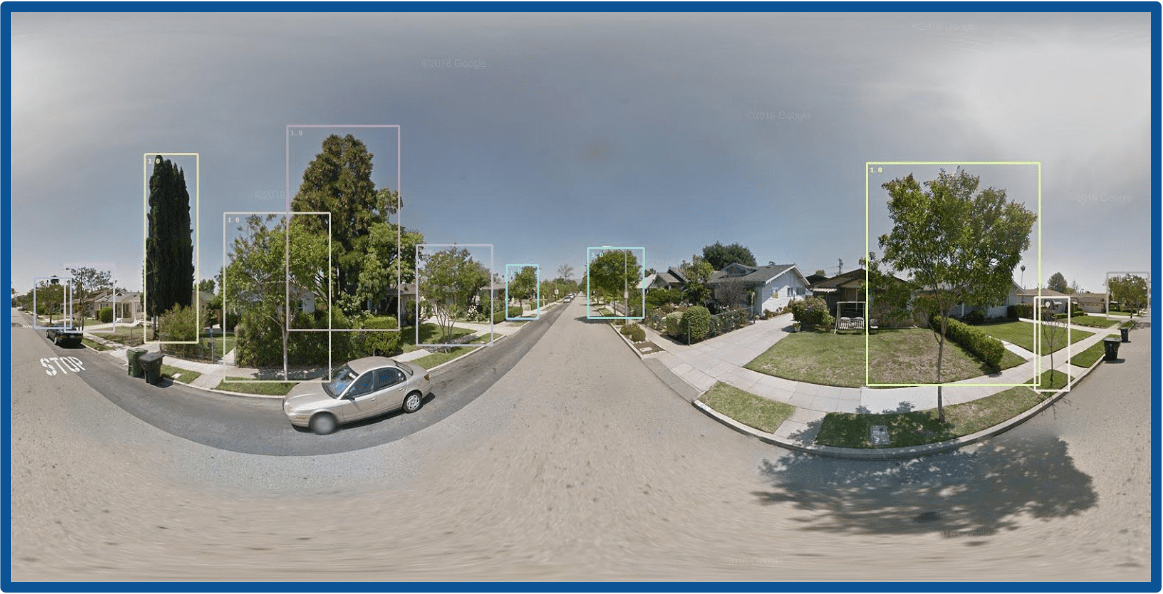}\\
		(c) & (d)\\
	\end{tabular}
	\Caption{Tree detection results of the example shown in Fig.~\ref{fig:nonmaxsuppr} projected back into the four individual street-view panoramas (color encodes camera). (Imagery \textcopyright~2019 Google)}
	\label{fig:nonmaxsuppr_sv}
\end{figure}

A positive side-effect of this procedure is that trees located far away from the street, usually located on private land and not part of the public street-tree population, are filtered out, too. Example results of this approach are shown overlaid on an aerial image in Fig.~\ref{fig:nonmaxsuppr} and projected back into the four individual street-view panoramas (color encodes camera) in Fig.~\ref{fig:nonmaxsuppr_sv}. 
\subsection{Matching detected trees to inventory entries}\label{method_treeassignment}

We cast the task of matching detected trees to tree entries in the database as an optimization problem where the total distance between all pairs of trees is minimized across an entire municipality. In order to cope with missing detections caused by, for example, occlusions in the image, missing street-view panoramas or simply missed trees by the detector, we set a maximum threshold of $M=50~meters$. In case the distance between a tree detected in the street-view panoramas and the geographic coordinate of the street address from a tree of the database exceeds this threshold, no tree is assigned. Our algorithm works as follows:
\begin{itemize}
	\item Given $n$ number of geocoded street addresses and $m$ number of geocoded detected trees, we compute the distance $dist_{i,j}$ between each address-tree pair ($i \in \{1,2,...,n\},~j \in \{1,2,...,m\}$).
	\item We define $x_{i,j}$ as a binary decision variable that indicates if an address $i$ is assigned to tree $j$. 
	\item Our objective function to be maximized is: $\max \sum_{i}\sum_{j}x_{i,j}\times (M-dist_{i,j})$ with M the distance threshold to avoid matching trees to database entries that are spaced too far apart.
	\item We introduce constraints to ensure that only one street address is assigned per tree $\sum_{j}x_{i,j}\leq1$ and that the number of assigned detected trees per street address does not exceed $K$, the total number of trees at that address given in the database $\sum_{i}x_{i,j}\leq K$.  
\end{itemize}
Although our approach strives for minimizing address-to-tree distances globally for an entire city, most tree assignment cases are locally constrained. This is also reflected by threshold M that discards pairs of distance larger then $50~meters$. However, distance computation between all possible pairs per data set is already very time consuming, needs a lot of memory, and neglects the local nature of the assignment problem. Therefore, we avoid explicit distance computation and compute a subset of tree candidate matches in the local vicinity of each street address via thresholding geographic latitude and longitude. This procedure leads to a sparse matrix representation of the assignment problem, where distances are only computed between a street address and detected trees inside a local region of interest. We finally minimize the discrepancy between all street addresses and all detected trees globally using Linear Programming. 

\section{Results and Discussion}\label{sec:experiments}

We ran experiments with tree inventories from five cities in the state of California, USA, that come with both accurate geocoordinates and street addresses, which allows us to train, validate, and test our approach. The total number of input trees with street addresses is 57938 and the number of street trees per city varied between 599 (Brentwood) and 34585 (Palo Alto). The total number of trees per municipality is reported in the top row of Tab.~\ref{tab:module_errors_numbers}. 

All street tree inventories were originally collected by different companies and stakeholders. This created a lot of inconsistent notation across the original data sets. In a first step, we thus preprocessed all tree inventory files with street addresses into the same format. Geo-coordinates of trees were originally recorded in the local Californian reference system CCS83 and needed to be transformed to WGS84, which we used for tree detections. 

\begin{figure}[!ht]
	\centering
	\includegraphics[width=0.8\linewidth]{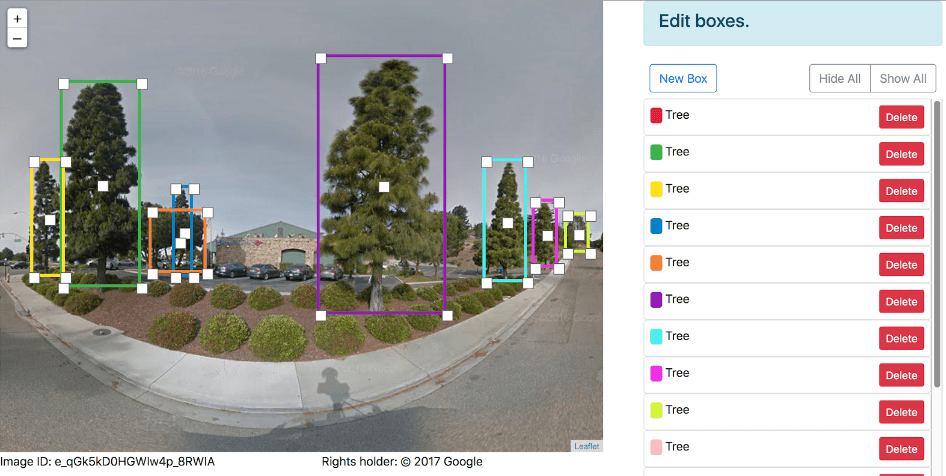} \\
	\Caption{Visualization of our labeling tool for manually drawing bounding boxes around individual trees in street-view panoramas where colors indicate different tree instances. (Imagery \textcopyright~2019 Google)}
	\label{fig:labeltool}
\end{figure}

In order to train the tree object detector~\citep{renNIPS15fasterrcnn}, we manually labeled 6783 individual tree instances in 718 street-view panoramas with a newly developed labeling tool (Fig.~\ref{fig:labeltool}). We sampled data locations for manual labeling by randomly selecting trees from inventories. All panoramas within a $10~m$ radius around the geographic position of a tree were downloaded for labeling, which resulted in one to four panoramas per selected tree. All trees in all panoramas were then labeled manually as shown in Fig.~\ref{fig:labeltool} with our labeling tool\footnote{The labeling tool will be made publicly available upon publication of this paper.}. 

\subsection{Tree geo-coding results}\label{eval_workflow}
\begin{sidewaysfigure}[!htbp]
	\centering
	\begin{tabular}{cc}
	\includegraphics[width=7.4cm]{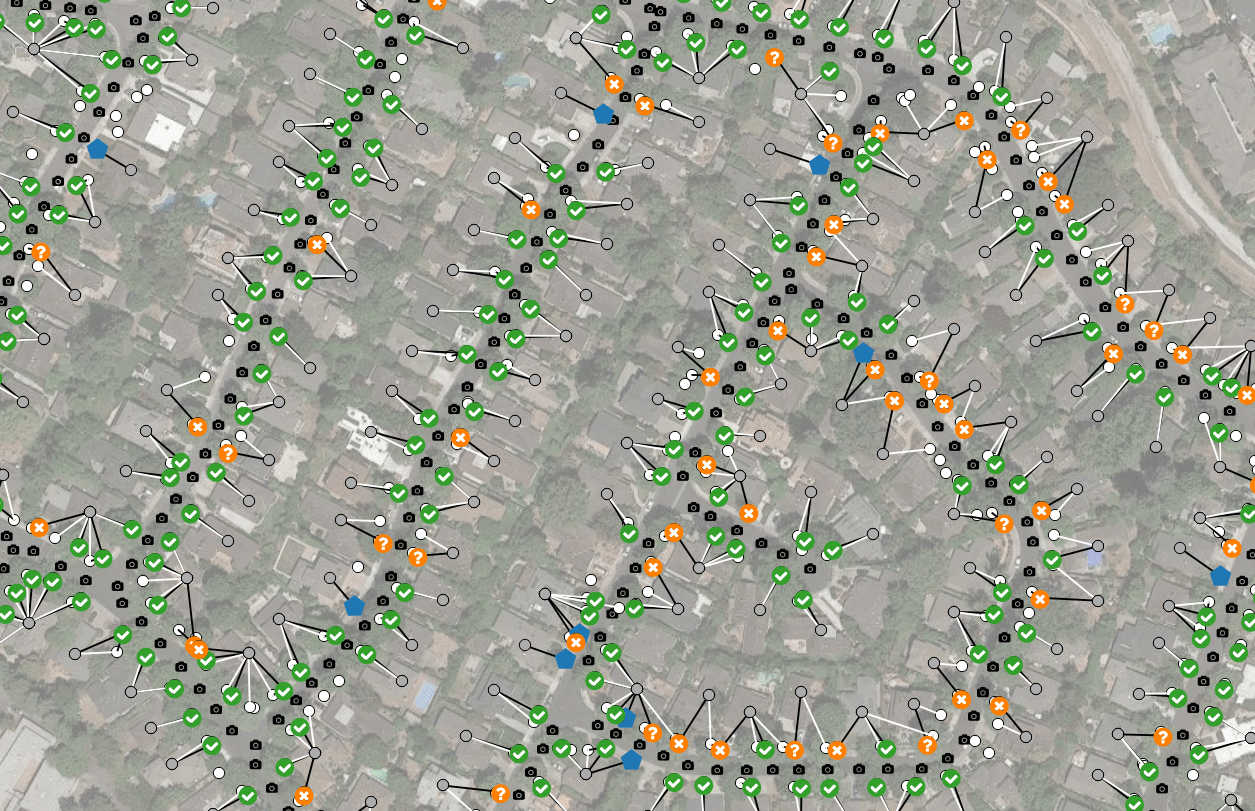} &
	\includegraphics[width=7.9cm]{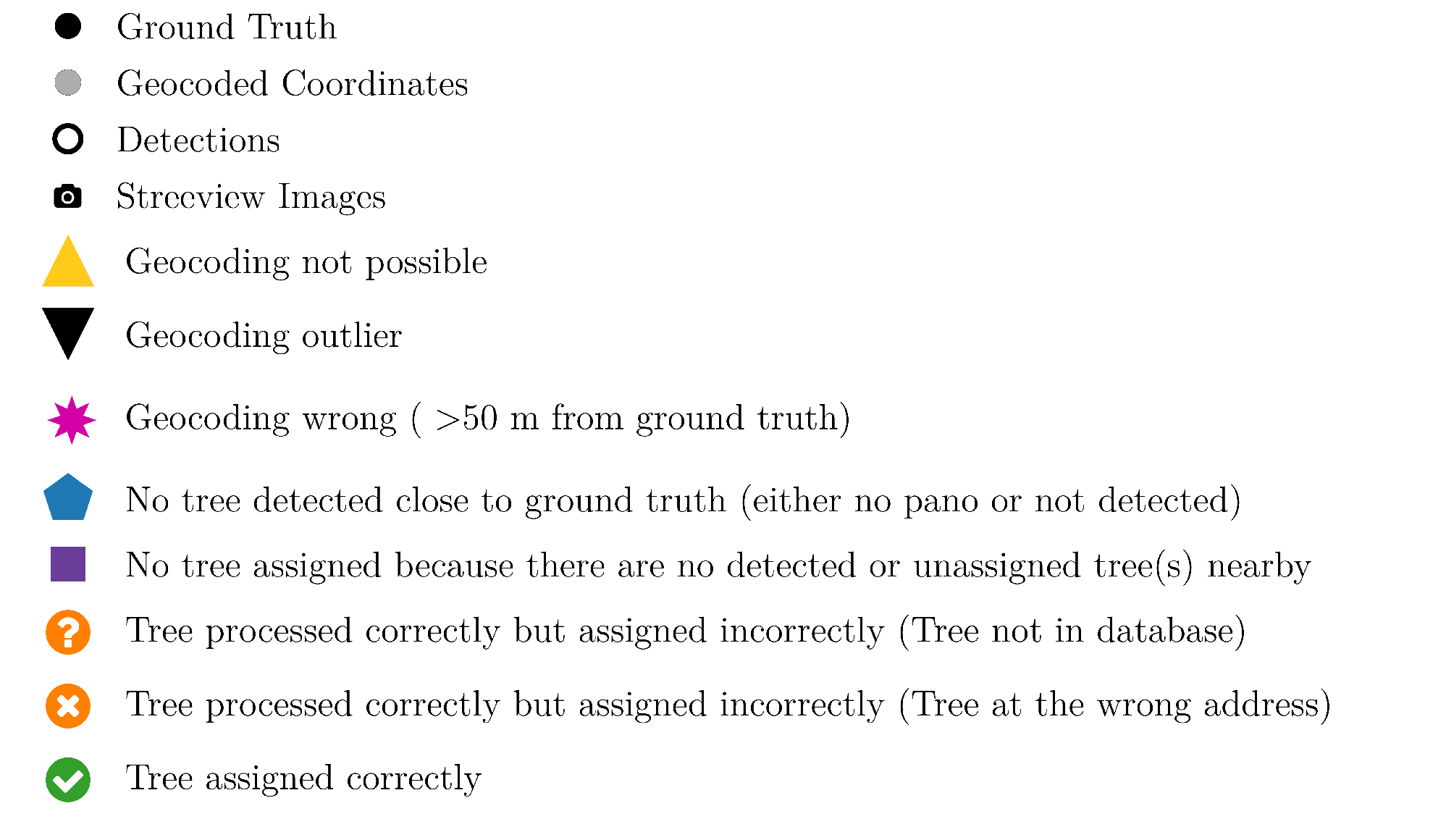} \\
	(a) & (b)\\
	\includegraphics[width=7.4cm]{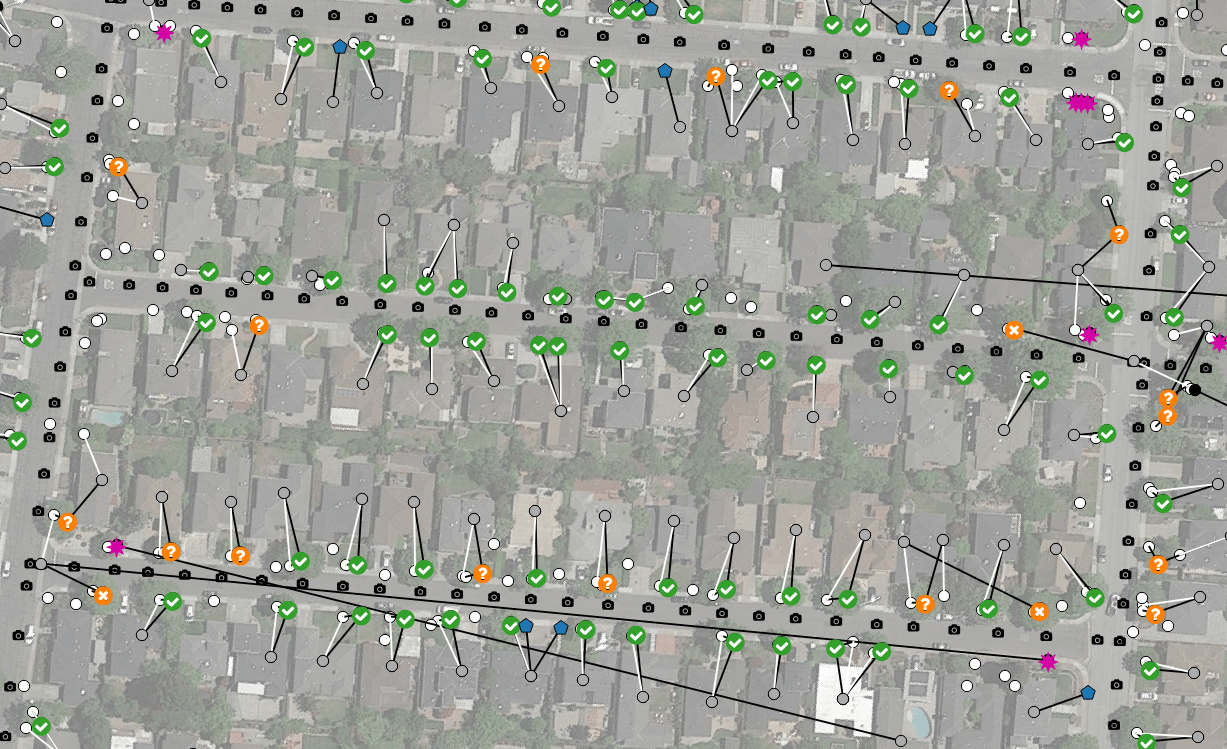} &
	\includegraphics[width=7.9cm]{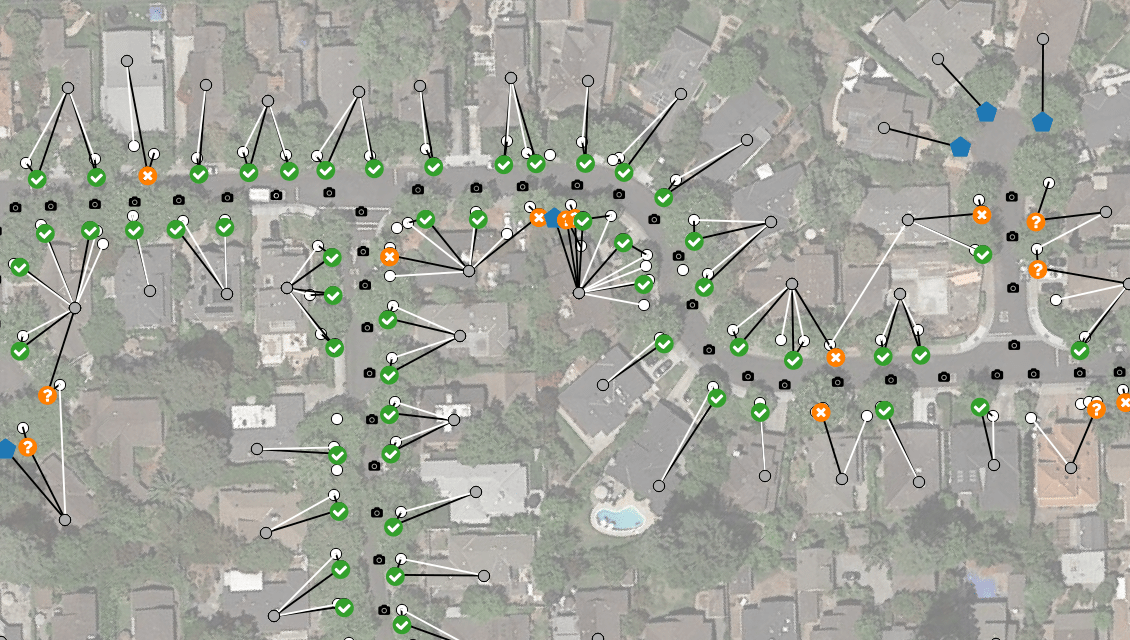} \\
	(c) & (d)\\
	\end{tabular}
	\Caption{(a,c,d) Results of the full processing pipeline overlaid on aerial images from Google Maps. (b) Legend of symbols used to represent correct cases and different error reasons. We can observe that most errors are trees that are processed correctly but assigned to a wrong address (orange dot with white x) as shown quantitatively in the second-to-last row ``Tree assigned incorrectly" of Tab.~\ref{tab:module_errors_numbers} (Imagery \textcopyright~2019 Google)}
	\label{fig:goodresults}
\end{sidewaysfigure}

\begin{figure}[!htbp]
\centering
\includegraphics[width=0.65\linewidth]{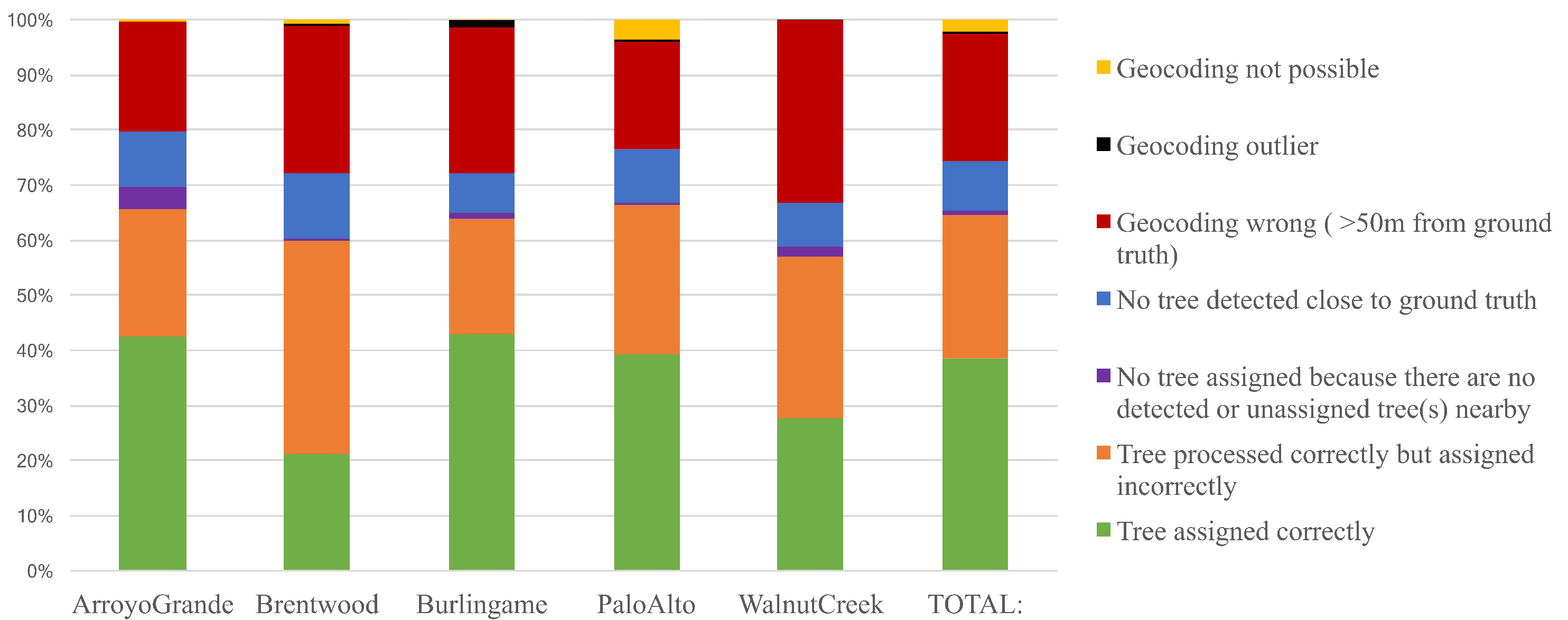} \\
\Caption{Module errors with respect to total number of inventory trees. The most dominant sources of error are wrong geocoding (red) and correctly processed trees that are assigned to the wrong address (orange) as shown quantitatively in Tab.~\ref{tab:module_errors_numbers}.}
\label{fig:module_errors}
\end{figure}

%
In this section, we present quantitative results for automatically assigning geographic coordinates to an inventoried tree. We also analyze and discuss failure cases in detail. In total, our approach correctly assigned geographic coordinates to 22363 individual street trees out of 57938 input trees with street addresses for five different municipalities where we have geographic ground truth positions. Results for the full processing pipeline are shown in Fig.~\ref{fig:goodresults} overlaid on aerial images of three sample sites. Quantitative results including reasons for errors are reported in Tab.~\ref{tab:module_errors_numbers} and visually shown as bar chart plots in Fig.~\ref{fig:module_errors}.   
\begin{sidewaystable}[!htbp]
\begin{center}
    \begin{tabular}{p{3cm}|p{1.5cm}|p{1.5cm}|p{1.5cm}|p{1.5cm}|p{1.5cm}|p{1.5cm}} 
 & Arroyo Grande &	Brent- wood &	Burlin- game &	Palo Alto &	Walnut Creek & TOTAL \\
\hline
\hline
\textbf{Tree number} &	\textbf{1701} & \textbf{599} & \textbf{13617}	& \textbf{34585}	& \textbf{7436}& \textbf{57938}	 \\
\hline
\hline
Geocoding\newline not possible&4\newline(0.2)&5\newline(0.8)&3\newline(0.0)&1209\newline(3.5)&0\newline(0.0)&1221\newline(2.1) \\
\hline
Geocoding\newline outlier&0\newline(0.0)&2\newline(0.3)&180\newline(1.3)&165\newline(0.5)&3\newline(0.0)&350\newline(0.6) \\
\hline
Geocoding\newline wrong &343\newline(20.2)&160\newline(26.7)&3600\newline(26.4)&6722\newline(19.4)&2474\newline(33.3)&13299\newline(23.0)\\
\hline
No tree\newline detected &170 (10.0)&71	(11.9)&987 (7.2)&3364 (9.7)&587 (7.9)&5179	(8.9) \\
\hline
No tree\newline assigned &65\newline(3.8)&2\newline(0.3)&131\newline(1.0)&159\newline(0.5)&128\newline(1.7)&485\newline(0.8) \\
\hline
Tree assigned\newline incorrectly &393\newline(23.1)&232\newline(38.7)&2886\newline(21.2)&9336\newline(27.0)&2194\newline(29.5)&15041\newline(26.0) \\
\hline
\textbf{Tree correct} &\textbf{726}\newline\textbf{(42.7)}&\textbf{127}\newline\textbf{(21.2)}&\textbf{5830}\newline\textbf{(42.8)}&\textbf{13630}\newline\textbf{(39.4)}&\textbf{2050}\newline\textbf{(27.6)}&\textbf{22363}\newline\textbf{(38.6)}\\
    \end{tabular}
 \end{center}
      \caption{Total number of errors per category and as percentage of the total number of inventory trees in each municipality (in parentheses)}
    \label{tab:module_errors_numbers}
\end{sidewaystable}
One reason for failure is impossible geocoding of a street address in the original inventory database (2.1\% of all cases; yellow triangle in Fig.~\ref{fig:goodresults}, yellow in bar plot Fig.~\ref{fig:module_errors}). This error occurs if the Google geocoding API cannot match a street address as presented in the original inventory database to a geographic coordinate, which is usually caused by incorrectly spelled or recorded addresses. A more severe failure case is incorrect geocoding of a given street address, i.e. a geographic position far away from the true position is assigned. We distinguish between two geocoding error types: geocoding outliers (0.6\% of all cases, black triangle in Fig.~\ref{fig:goodresults}, black in bar plot Fig.~\ref{fig:module_errors}) and geocoding farther then 50 meters from the true position (23.0\% of all cases, purple star in Fig.~\ref{fig:goodresults}, red in bar plot Fig.~\ref{fig:module_errors}). Geocoding outliers are cases where a street address is projected far away from all other entries in the database, as detected via the z-score, described in Sec.~\ref{sec:geocoding}. Reasons include incorrectly spelled or recorded addresses. Geocoding that deviates more than 50 meters from the true position can also be caused by incorrect spelling or recording of a street address. Another reason is the way the Google geocoding API works. Most street addresses ($\approx68\%$ of all cases) are geocoded via rooftop coordinates of the center of the main building at a given street address. If these are unavailable, the API estimates a geographic coordinate. This is done by either interpolating between two precise points like road intersections ($\approx27\%$) or, if impossible, by returning the geometric center of a street or a known region ($\approx5\%$).    

Another case is if no tree is detected at a given inventory tree location, either due to missing streetview panorama or a missed detection (8.9\% of all cases, blue pentagon in Fig.~\ref{fig:goodresults}, blue in bar plot Fig.~\ref{fig:module_errors}). Furthermore, it can happen that all detected trees in the vicinity of an inventory tree have been assigned to neighboring tree entries during the assignment step (0.8\% of all cases, purple square in Fig.~\ref{fig:goodresults}, purple in bar plot Fig.~\ref{fig:module_errors}).

Finally, when detected trees are matched to database entries, there are erroneous assignments. 26.0\% of all inventory database entries received incorrect tree assignments (orange in barplot Fig.~\ref{fig:module_errors}). To determine if an assignment is correct, we check if a detected tree is matched to the correct geocoded address. Note that we could not check if each individual tree was matched correctly in case of multiple trees per address (83\% of all trees share a street address with more trees). Although the database contains information about tree species and height, too, which would in theory allow to distinguish neighboring trees, the vast majority of neighboring trees turned out to be of the same species with the same height.  

There are several reasons for assigning a wrong tree to an entry in the database (orange in bar plot Fig.~\ref{fig:module_errors}). Since our assignment optimizer rewards spatial proximity between detected tree and geocoded street address of the inventory tree entry, a tree close to the boundary between two parcels can be assigned to the wrong one (orange dot with white x in Fig.~\ref{fig:goodresults}). Errors can occur as a result of detected trees that were originally not recorded in the database (orange dot with white question mark in Fig.~\ref{fig:goodresults}). This can happen because the trees were originally missed during the field campaign, a new tree was planted, or a tree on private land is detected (i.e., and thus was not recorded in the inventory of public street-trees). If these trees are matched to an entry in the inventory database, they also count as incorrect assignments.

\subsection{Error analysis}\label{sec:discussion}

In this section we analyse in detail the dominant reasons for failure. We look at the influence of geocoding quality, erroneous or missing tree detections, and the assignment step.

\begin{figure}[!htbp]
	\centering
	\includegraphics[width=0.95\linewidth]{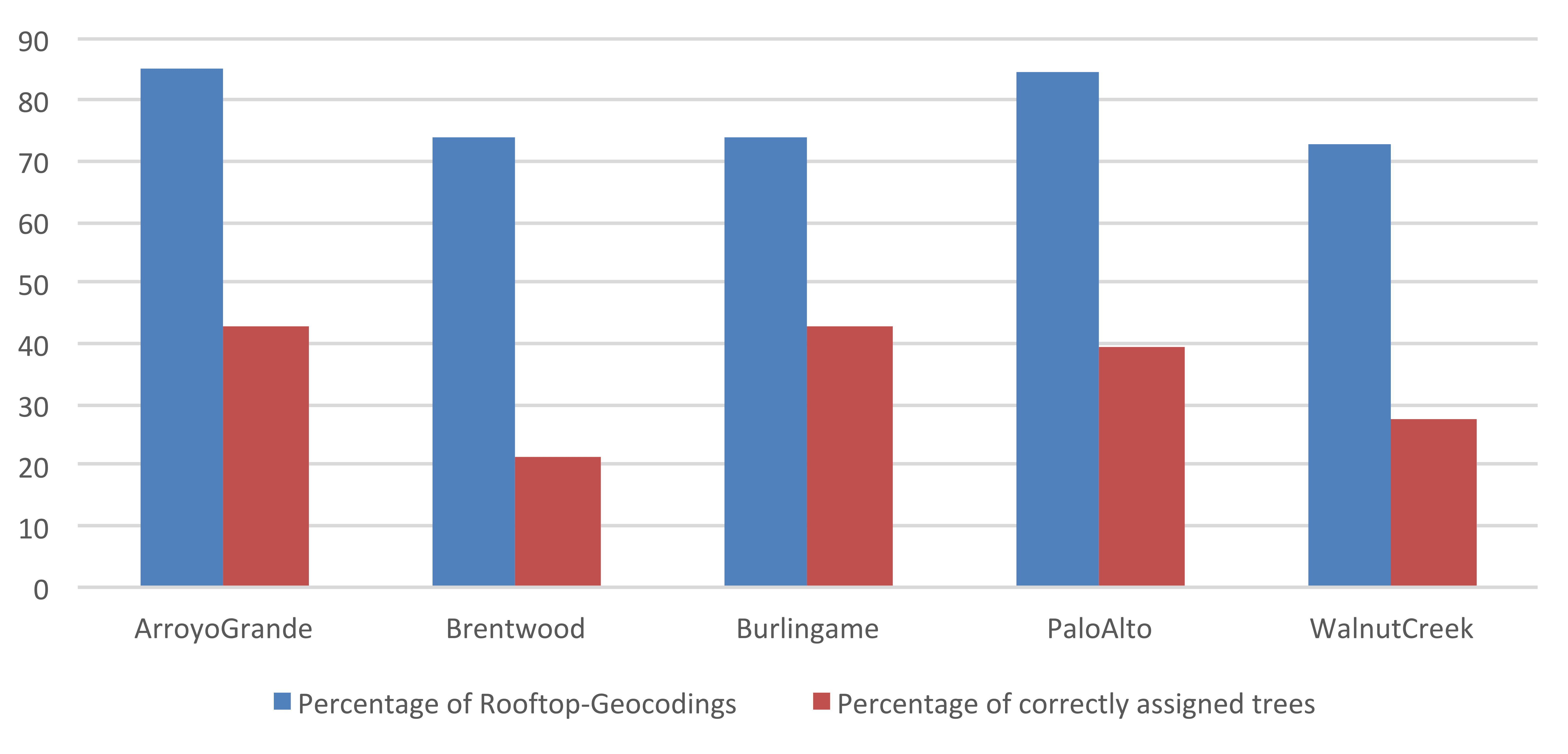}\\
	\Caption{Comparison between geocoding via rooftop coordinates and correctly assigned trees. Higher percentages of rooftop-geocodings (blue) lead to higher percentages of correctly assigned trees (red).}
	\label{fig:compare_GT}
\end{figure}

\begin{sidewaysfigure}[!htbp]
	\centering
	\begin{tabular}{cc}
		\includegraphics[height=7cm]{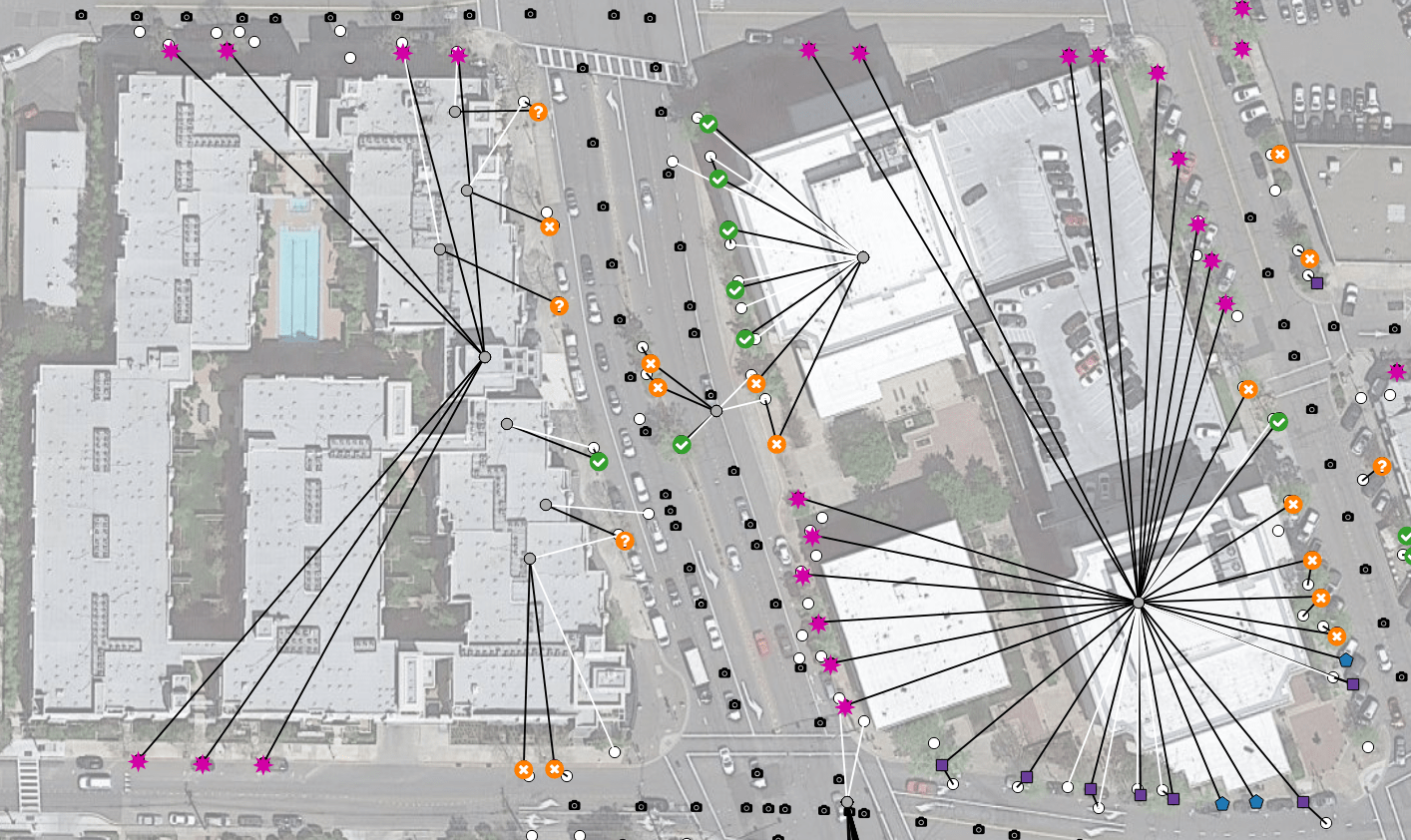}& 
		\includegraphics[height=7cm]{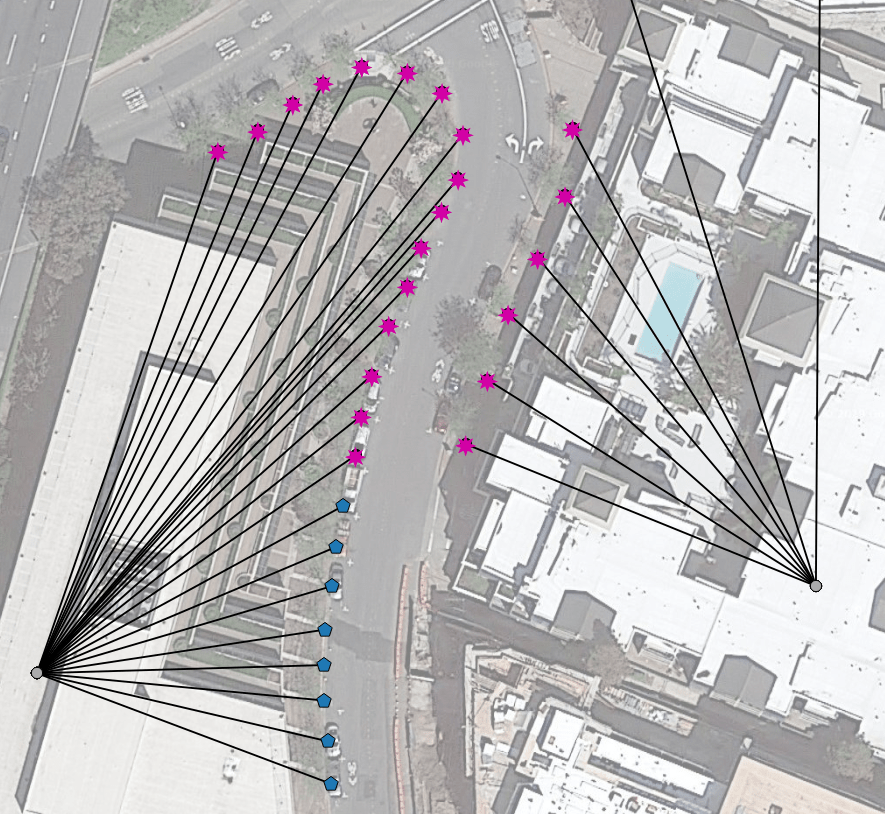}\\
		(a) & (b)\\
	\end{tabular}
	\Caption{Examples for failure of geocoding (rose stars, refer to legend in Fig.~\ref{fig:goodresults}(b)): (a) and (b) show street addresses of large buildings with a high number of trees. However, most detected trees are farther away than $50~meters$ (rose star) from the rooftop coordinate and thus count as unassigned. Note that many of these cases are still located close to a tree in the inventory database (white dot) and could be manually processed. (Imagery \textcopyright~2019 Google)}
	\label{fig:treegeocodingrrors}
\end{sidewaysfigure}

\textbf{Geocoding quality} has a big impact on the number of correctly assigned trees. The best possible kind of geocoding are rooftop coordinates in the center of the main building per address as compared to more approximate methods such as the ones described earlier that for example interpolate between road intersections. We plot the proportion of rooftop coordinates per municipality versus the proportion of correctly assigned trees in Fig.~\ref{fig:compare_GT}. In general, a high proportion of successfull street address geocoding via rooftop coordinates usually leads to higher numbers of correctly assigned trees. Palo Alto and Arroyo Grande have the highest proportion of rooftop coordinates with $\approx88\%$ and the lowest proportion of geocoding errors (19.4\% and 20.2\%). Burlingame looks like an exception at first glance, having a low number of rooftop coordinates, but highest number of correctly assigned trees (42.8\%). However, a look into Tab.~\ref{tab:module_errors_numbers} reveals that the geocoding error was relatively high (26.4\%).
In contrast, large parcels with very large buildings like the one shown in Fig.~\ref{fig:treegeocodingrrors} led to a very high amount of errors for several reasons. Large parcels (and buildings) led to large distances between street-trees and rooftop coordinates. Consequently, geocoding is evaluated incorrectly because the rooftop coordinate is farther than 50 meters from the tree position at that street address (purple star in Fig.~\ref{fig:treegeocodingrrors}).  
%



\begin{sidewaysfigure}[!htbp]
	\centering
	\begin{tabular}{cc}
		\includegraphics[height=7cm]{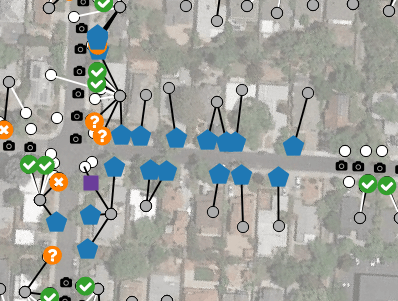}& 
		\includegraphics[height=7cm]{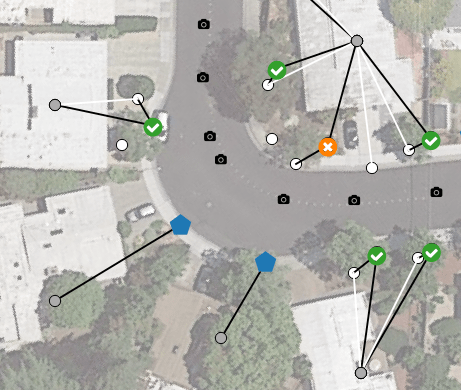}\\
		(a) & (b)\\
	\end{tabular}
	\Caption{Examples for typical failure cases of tree detection in panorama images (refer to legend in Fig.~\ref{fig:goodresults}(b)): (a) no tree detected in the street-view panorama (blue pentagons), (b) no street-view panorama available at junction leading to undetected trees (blue pentagons). (Imagery \textcopyright~2019 Google)}
	\label{fig:treedetectionerrors2}
\end{sidewaysfigure}

\begin{sidewaysfigure}[!ht]
	\centering
	\begin{tabular}{ccc}
		\includegraphics[height=4.5cm]{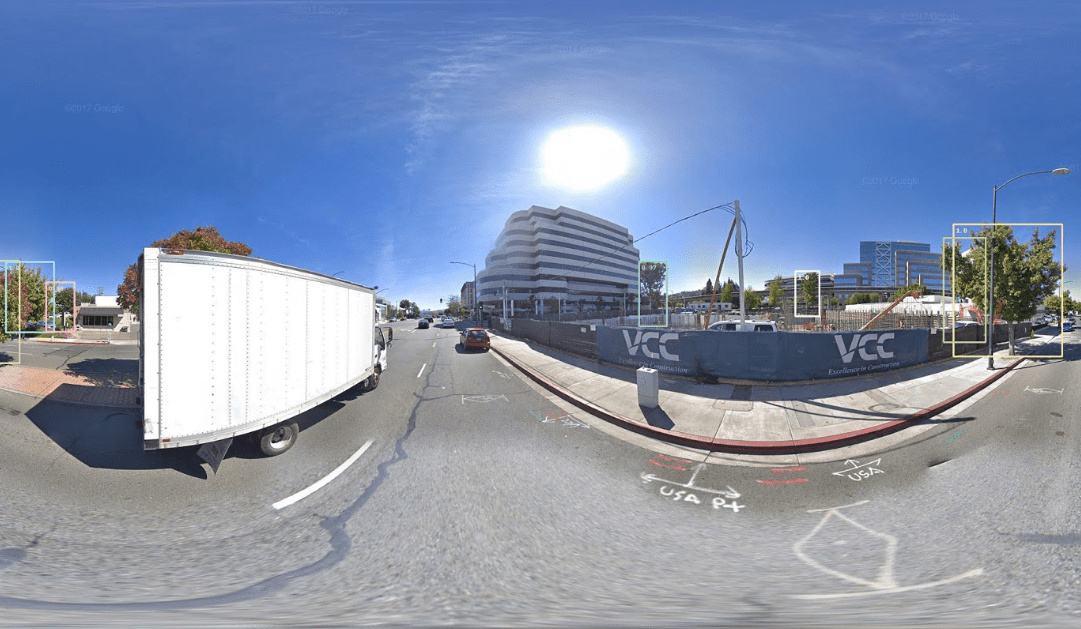}& 
		\includegraphics[height=4.5cm]{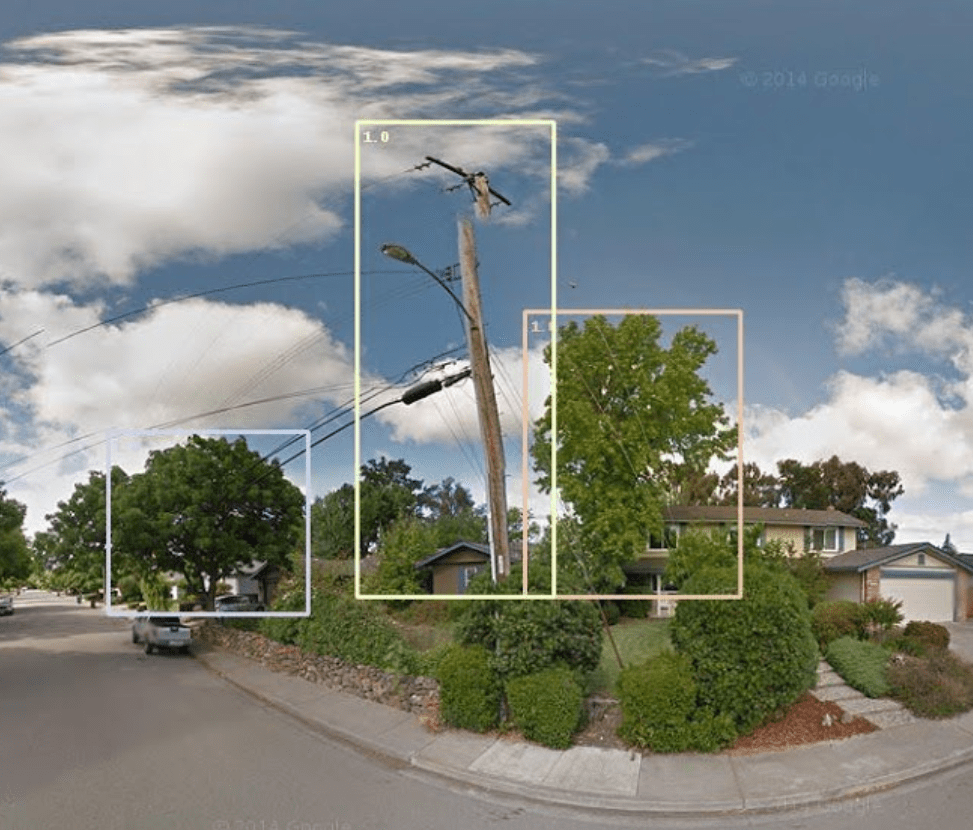}& 
		\includegraphics[height=4.5cm]{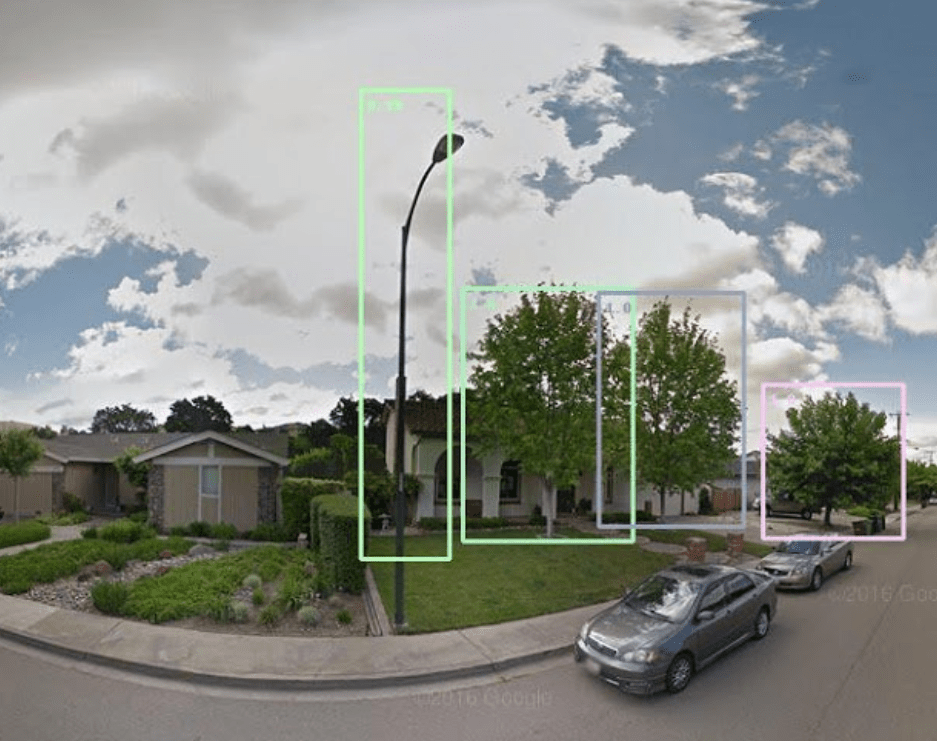}\\
		(a) & (b) & (c)\\
	\end{tabular}
	\Caption{False positives and false negatives of tree detection: (a) tree occluded by a truck, (b) power-pole wrongly classified as tree, (c) street lamp wrongly classified as tree. (Imagery \textcopyright~2019 Google)}
	\label{fig:treedetectionerrors1}
\end{sidewaysfigure}

\textbf{Tree detections} in Google street-view panoramas can sometimes also fail either due to missing panoramas or an error by the detector. Tree detection failure caused 8.9\% of all entries in the inventory database to remain without a valid geo-coordinate (see category \textit{no tree detected} in Tab.~\ref{tab:module_errors_numbers}). One major reason is that at various locations no street-view panorama was acquired close enough to the tree position. An example scene is shown in Fig.~\ref{fig:treedetectionerrors2}(a), where black camera symbols indicate acquisition locations of panoramas and blue pentagons indicate database entries without any detected tree. In case panoramas are available, the detector may sometimes miss a tree as is shown with the two blue pentagons in Fig.~\ref{fig:treedetectionerrors2}(b). The dominant reason for missed trees (i.e., false negatives) are occlusions. If trees are partially or completely occluded in the panorama by, for example, a truck (Fig.~\ref{fig:treedetectionerrors1}(a)), there is no way to recover them and assign a coordinate to a database entry at that address. Another example of a false negative is if a tree in the field is removed between the time that it was inventoried and the time that a street-view image is recorded so the API does not detect a tree but it used to be there. Two examples of false positives, i.e. objects falsely detected as trees, are shown in Fig.~\ref{fig:treedetectionerrors1}(b,c). A more detailed analysis of detection error cases and proportions is given in \citet{branson2018}. Note these false detections in individual images usually do not survive the local non-maximum suppression step of our workflow because they are rarely detected in more than one panorama.
%


%
%
\textbf{Trees assigned incorrectly} are another major reason for mismatches. Incorrect assignments of detected trees accounted for 26.0\% of all entries in the database that were matched incorrectly. Quantitative results in Tab.~\ref{tab:module_errors_numbers} show that the error distribution and also the total number of successfully assigned trees varied across different municipalities. For example, Brentwood with the lowest number of public street-tree entries in the inventory compared to the four other municipalities, had also the lowest number of correctly assigned trees (21.2.\%). The error distribution reveals that a very high percentage of trees is incorrectly assigned. A closer look at Brentwood shows that it is a suburban municipality of relatively low density with many trees close to the street on private land and only very few, scarcely distributed public street-trees. This leads to a high number of cases where trees on private land are wrongly matched to a geocoded inventory entry, often because the tree on private ground is closer to the geographic rooftop coordinate of the main building on a parcel. On the contrary, Burlingame with the highest percentage of assigned trees (42.8\%) is a municipality with denser suburban structure and many street-trees on public land that densely line most streets. Incorrectly assigned trees make up only 21.2\% of all of Burlingame's inventory trees as compared to Brentwood's 38.7\%.     


\subsection{Test on 48 municipalities and discussion}
\begin{figure}[!htbp]
\centering
\includegraphics[width=0.85\linewidth]{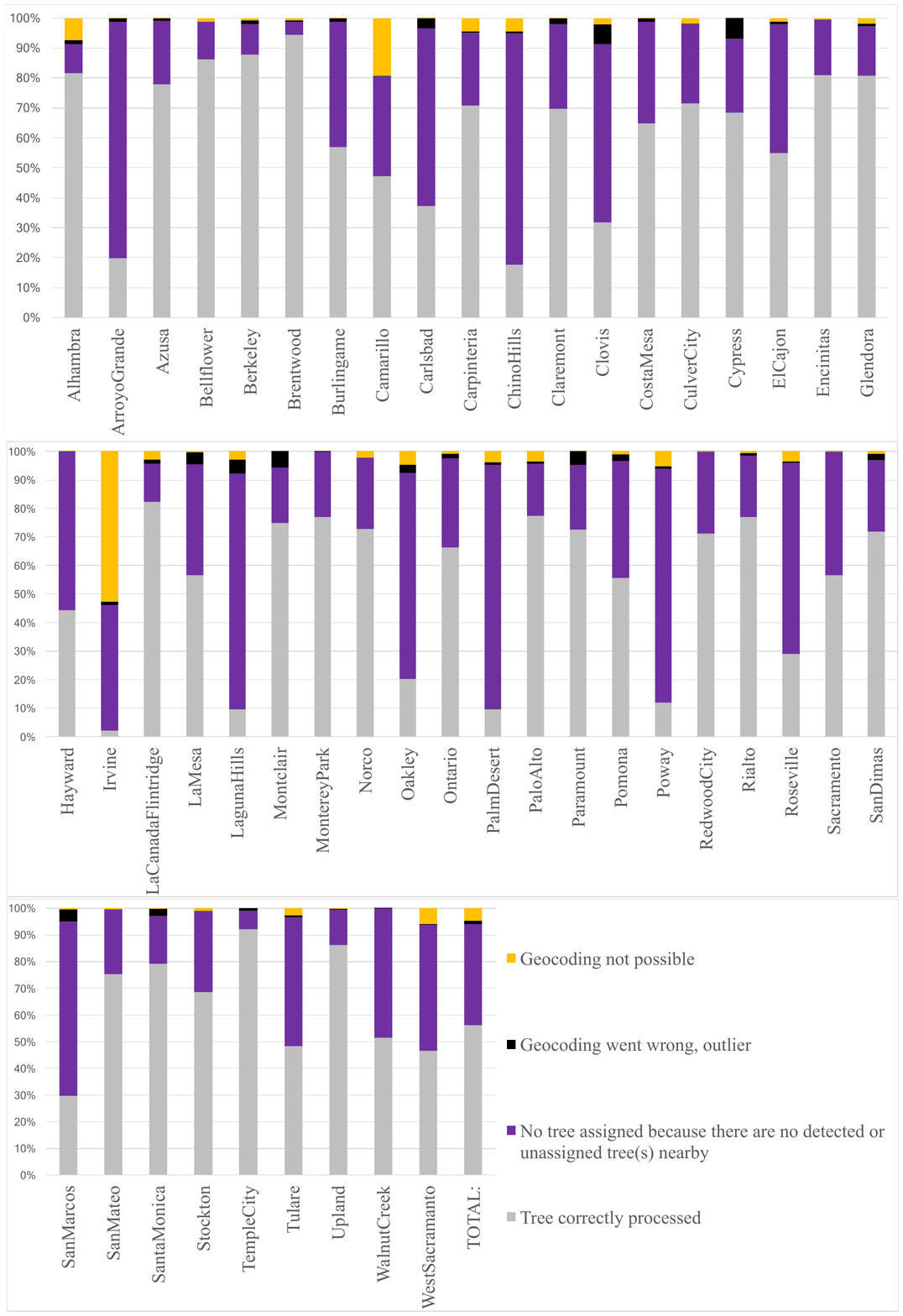} \\
\Caption{Results for all 48 cities containing in total 1100952 input trees from inventories. Error proportions in each municipality are color-coded by error type. Missing (detected) trees at a geocoded street address are the most dominant reason for errors (purple).}
\label{fig:module_errors_all}
\end{figure}
We ran our processing pipeline on all 48 municipalities containing 1100952 input trees across all inventories. The five municipalities that we used for a detailed error analysis because they have geographic coordinates in addition to street addresses, are a subset of these 48 municipalities. We show all results with errors at test time indicated in color in Fig.~\ref{fig:module_errors_all}. Our method assigned geographic coordinates to 619229 individual street trees (grey in Fig.~\ref{fig:module_errors_all}), which amounted to 56\% of all inventory trees. Note that error evaluation at test time (i.e., geographic ground truth position per tree available) can only detect the following failure cases (as shown in Fig.~\ref{fig:module_errors_all}):
\begin{itemize}
    \item geocoding of the street address is impossible (yellow),
    \item or generates an outlier according to our z-score (black),
    \item and missing tree detections in the panorama images close to the geocoded street address of a tree in the inventory (purple).
\end{itemize} 
Since we do not have ground truth tree coordinates in this realistic application scenario, we cannot detect all possible failure cases. More precisely, we cannot recognize cases where geocoding is wrong, i.e. geocoding of a given street address passes the z-score but is more than 50 meters away from the true tree position, and where trees are assigned incorrectly. Major reasons for incorrect geocoding are approximate geocoding (no rooftop coordinates, but interpolation between road intersections, for example) and very large parcels and buildings with a very high number of street-trees (Fig.~\ref{fig:treegeocodingrrors}). A simple way to significantly reduce the amount of errors caused by erroneous geocoding is filtering out all trees matched via non-rooftop geocoding. We evaluated this idea on those five municipalities where we have ground truth coordinates (and assume a similar behaviour for all other municipalities without ground truth coordinates). Elimination of all cases without rooftop coordinates decreases the total percentage of incorrect geocoding across the five municipalities by 11.4 percentage points to 11.6\%. The overall correctly assigned tree percentage increases by 8.4 percentage points to 47\% (see Tab.~\ref{tab:module_errors_numbers}).

Incorrect assignments of trees due to mixing up trees at the boundaries between two parcels or confusion between trees on private land and street-trees is generally difficult to fix. One direction for automatic detection of these errors would be to train a tree species classifier as was done in \citet{branson2018} to verify whether the species of the tree in the database does actually match the tree in the images. While this would probably work well to detect cases of confusion between trees on private lawns and street-trees (often different species), it would rarely solve mix-ups between two neighboring street-trees at parcel boundaries because these often have the same species. 



%


\section{Conclusion}\label{sec:conclusion}

We have presented a novel approach to assign geographic coordinates to street-trees given street addresses and street-view panorama images. Enhancing existing tree inventories with geographical locations allows to automatically match inventories from different dates and to track trees across longer time spans. Depending on the information that is being recorded in tree inventories (e.g., presence/absence, trunk diameter, vigor), repeated measures make it possible to calculate tree mortality rates and detect changes in tree health. Linking legacy inventories to newer inventories is a cost- and time-effective way to obtain a much larger sample size than is typically available to municipalities and researchers. Urban tree demographic studies have for the most part based their analyses on a small number of samples limited in geographic area. Our hope is that with the proposed approach, the sample of the cities' tree population can be increased tremendously. 

Experimental evaluation of our method on five different municipalities with geographic coordinates show that the method is capable of assigning geographic coordinates correctly to 38.6\% of the trees in the inventory databases. Application to all 48 Californian municipalities with inventory data demonstrates that we can assign geographic locations to 619229 trees out of 1100952 trees (56\%). After a simple post-processing step that filters out all addresses geocoded via less accurate, non-rooftop coordinates, geocoding errors can be reduced by $\approx11$ percentage points and the overall percentage of correctly assigned trees can be increased by $\approx8$ percentage points. 

The major shortcoming of the proposed method is that 26\% of trees were assigned incorrectly (Tab.~\ref{tab:module_errors_numbers}), which is an error that can hardly be detected during a realistic scenario without available ground truthed tree coordinates. Our future work will thus look into strategies to significantly reduce this type of error by implementing strategies such as adding tree species information to the matching step.

We hope this work encourages further research to generate longer time series (over several decades) of street-tree inventories with the ultimate goal of better understanding street tree population dynamics and corresponding changes in ecosystem services at a very large scale.

\section{Acknowledgements}\label{sec:ackn}
This project was supported by funding from the Hasler Foundation, the US Department of Agriculture-Forest Service, and the Swiss National Science Foundation scientific exchange grant IZSEZ0\_185641.

{
	  \bibliographystyle{elsarticle-harv}
		\bibliography{IJPRS_address2geo} 
}

\newpage
\appendix

\end{document}